%% file: main.tex
\documentclass[letterpaper, 10 pt, conference]{ieeeconf}
\IEEEoverridecommandlockouts

\usepackage{times}
\usepackage[bookmarks=true]{hyperref}

\input{preamble.tex}

\input{commands.tex}
\newcommand{\citet}[1]{\cite{#1}}
\newcommand{\citep}[1]{\cite{#1}}

\begin{document}

\title{\titlelong}

\author{
Yves-Simon Zeulner,
Simon Crämer,
Sandeep Selvaraj,
Roberto Calandra%
\thanks{All authors are with LASR Lab, TU Dresden, Dresden, Germany}%
\thanks{Corresponding author: \href{mailto:yves-simon.zeulner@mailbox.tu-dresden.de}{yves-simon.zeulner@mailbox.tu-dresden.de}}%
\thanks{Helmholtzstr. 18, 01069 Dresden, Germany, +49 351 463-40850}
}

\maketitle

\begin{abstract}
\input{0_abstract}
\end{abstract}

\section{Introduction}
\input{1_introduction} 

\section{Related work}
\input{2_related}

\section{Learning to Play Piano in the Real World}
\label{sec:approach}
\input{3_approach}

\section{Experimental Results}
\input{4_result}

\section{Limitations and Future Work}
\input{5_limitations}

\section{Conclusion} 
\label{sec:conclusion}
\input{6_conclusion}

\section*{Acknowledgments}
\input{99_acknowledgments}

\bibliographystyle{IEEEtran}
\bibliography{references}

\section*{Appendix}
\input{appendix/appendix_main}

\end{document}

%% file: preamble.tex
\usepackage[english]{babel}

\usepackage{graphicx}
\usepackage{animate}
\usepackage{epsfig}
\usepackage{epstopdf}

\usepackage{listings}
\usepackage{color}
\usepackage{nameref}
\usepackage{hyperref}
\usepackage{amsmath}
\usepackage{amssymb}

\usepackage{dsfont}
\usepackage{mathtools}

\usepackage{epigraph}
\usepackage{lscape}
\usepackage[]{nomencl}
\usepackage{algorithm}

\usepackage{algorithmic}
\usepackage{multicol}
\usepackage{multirow}
\usepackage{etoolbox}

\usepackage{caption}
\usepackage{subcaption}
\usepackage{wrapfig}

\usepackage{siunitx}

\usepackage{floatflt}

\usepackage{url}

\usepackage{todonotes}

%% file: commands.tex
\newcommand{\eg}[0]{e.g.}

\newcommand{\email}[1]{\href{mailto:#1}{\nolinkurl{#1}}}

\renewcommand{\sec}[1]{Section~\ref{#1}}
\newcommand{\fig}[1]{Fig.~\ref{#1}}

\newcommand{\app}[1]{Appendix~\ref{#1}}
\newcommand{\tab}[1]{Table~\ref{#1}}

\newcommand{\titlelong}[0]{Learning to Play Piano in the Real World}
\newcommand{\website}[0]{\url{www.lasr.org/research/learning-to-play-piano}}

%% file: 0_abstract.tex
% Max 200 words
Towards the grand challenge of achieving human-level manipulation in robots, playing piano is a compelling testbed that requires strategic, precise, and flowing movements. 
Over the years, several works demonstrated hand-designed controllers on real world piano playing, while other works evaluated robot learning approaches on simulated piano playing.
In this work, we develop the first piano playing robotic system that makes use of learning approaches while also being deployed on a real world dexterous robot.
Specifically, we use a Sim2Real2Sim approach where we iteratively alternate between training policies in simulation, deploying the policies in the real world, and use the collected real world data to update the parameters of the simulator.
Using this approach we demonstrate that the robot can learn to play several piano pieces (including \textit{Are You Sleeping}, \textit{Happy Birthday}, \textit{Ode To Joy}, and \textit{Twinkle Twinkle Little Star}) in the real world accurately, reaching an average F1-score of $0.881$.
By providing this proof-of-concept, we want to encourage the community to adopt piano playing as a compelling benchmark towards human-level manipulation in the real world.
We open-source our code and show additional videos at \website{}.

% 0.8817933830878294

%% file: 1_introduction.tex
    
Playing the piano requires humans to master contact-rich hand movements dictated by the timing and tone they intend to produce. 
This mastery is not learned quickly but through extensive practice, which requires humans to control their hands based on the haptic and auditory feedback received with each key pressed on the piano. 
In addition, human hands are an extraordinary research subject due to their unmatched dexterity, precision, and adaptability. 
These capabilities have evolved to perform various tasks, from delicate manipulations like sewing to powerful tasks like shot-put. 
This makes human-like hands a natural inspiration for robotic systems.
In robotics, much attention has been paid to tasks such as in-hand object rotation, where the goal is to develop and optimize object manipulation through intricate movements \cite{Handa2024, wang2024pens, openai2019, Qi2022}. 
These studies often achieve efficient manipulation by exploiting unnatural and sometimes non-anthropomorphic motions to improve control and precision. 
Although such techniques yield impressive results, they often diverge from the natural movements of the human hand, which could limit their applicability to tasks that require more human-like motions. 
This stands in strong contrast to the task of playing a musical instrument like the piano. 
While piano playing demands a high level of motor control, coordination, and precision it also requires close alignment with the natural movements of human hands.

\begin{figure}[t]
    \centering
    \includegraphics[width=\linewidth]{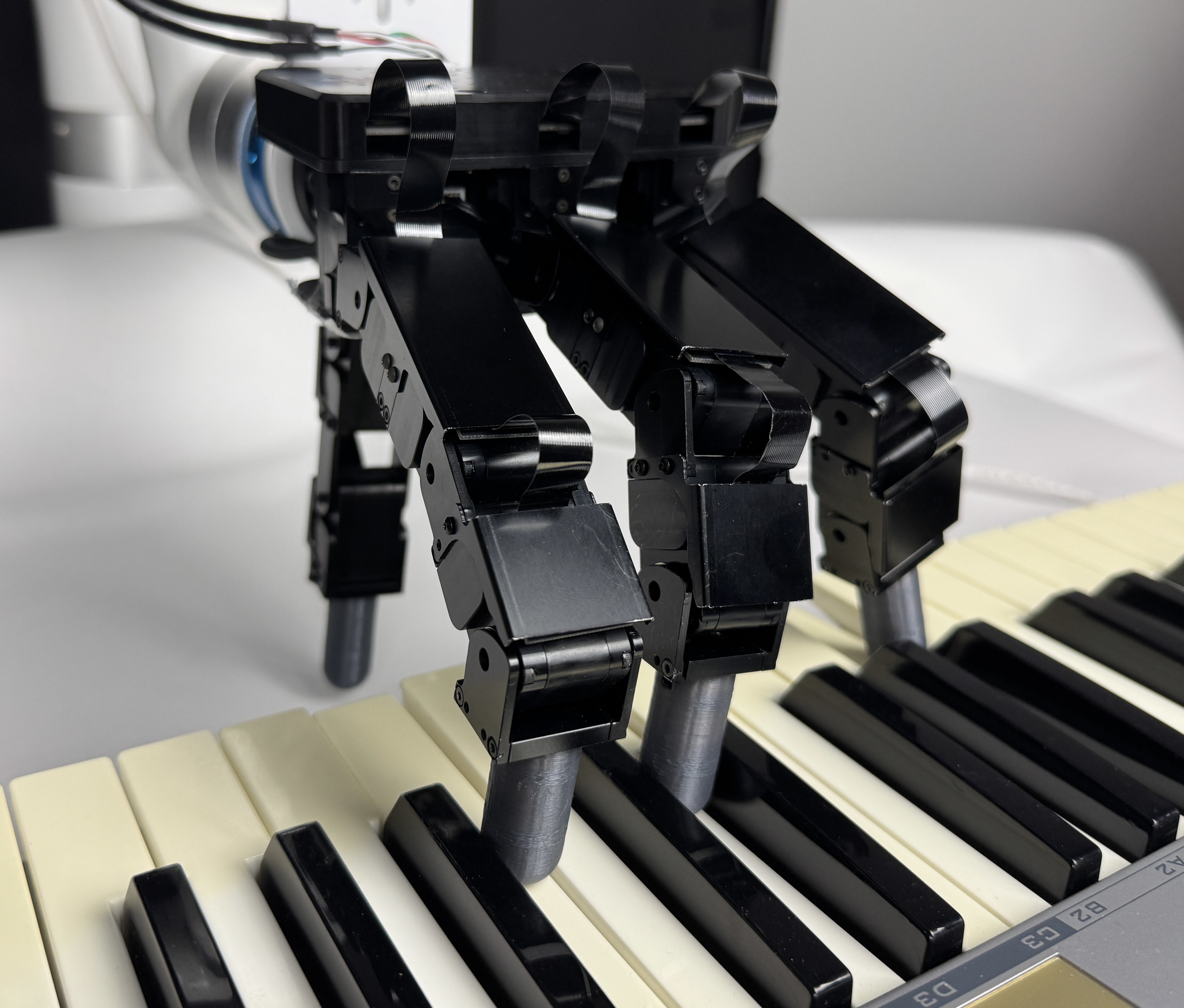}
    \caption{In this work, we demonstrate a proof-of-concept for learning to play piano with a real world robot. 
    To achieve this, we employed a multi-finger robot hand and a Sim2Real2Sim approach. 
    Experimental results show that the robot can learn to play several simple pieces successfully, after training exclusively in simulation.}
    \label{fig:teaser}
\end{figure}

Multiple works already explored RL agents for playing piano in simulation~\citep{Xu2022, Zakka2023, qian2024, Zhao2024, chen2025, huang2025}.
However, we are the first to deploy a learned piano playing agent on a physical robot. 
Our contribution focuses on transferring the learned strategies in simulation into the real world.
This is a demanding task due to the inaccuracies within the training environments and the real world (i.e., Sim2Real gap).
This Sim2Real gap can be reduced by tuning the simulation parameters to mimic the real world as close as possible and by training a robust policy.
Our approach uses an iterative Sim2Real2Sim approach that utilizes observations of the real world to optimize the training environment.
Additionally we use domain randomization during the training to make our learned policy more robust.
This enables us to play recognizable songs (without any modification of the tablatures), such as \textit{Are You Sleeping}, \textit{Happy Birthday}, \textit{Ode To Joy}, and \textit{Twinkle Twinkle Little Star}, with a multi-finger Allegro hand successfully on a real world piano. 
By providing this proof-of-concept, we want to encourage the community to adopt piano playing as a compelling benchmark towards human-level manipulation.

\begin{figure*}[t]
\begin{subfigure}[t]{0.49\linewidth}
 \centering
  \includegraphics[width=1\linewidth]{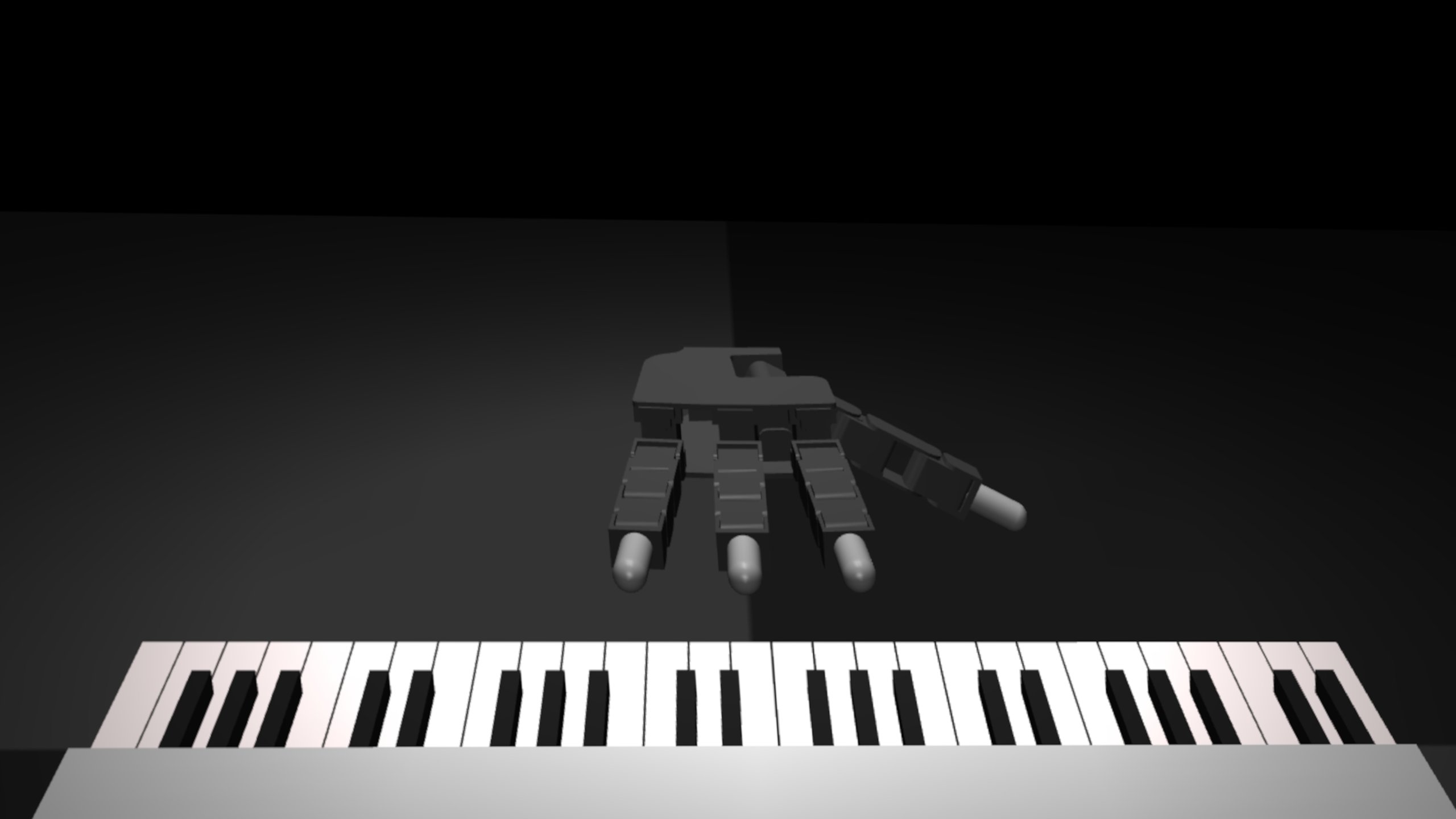}
  \caption{Simulated hand and piano.}\label{fig:sim env}
\end{subfigure}
\hfill
\begin{subfigure}[t]{0.49\linewidth}
  \centering
  \includegraphics[width=1\linewidth]{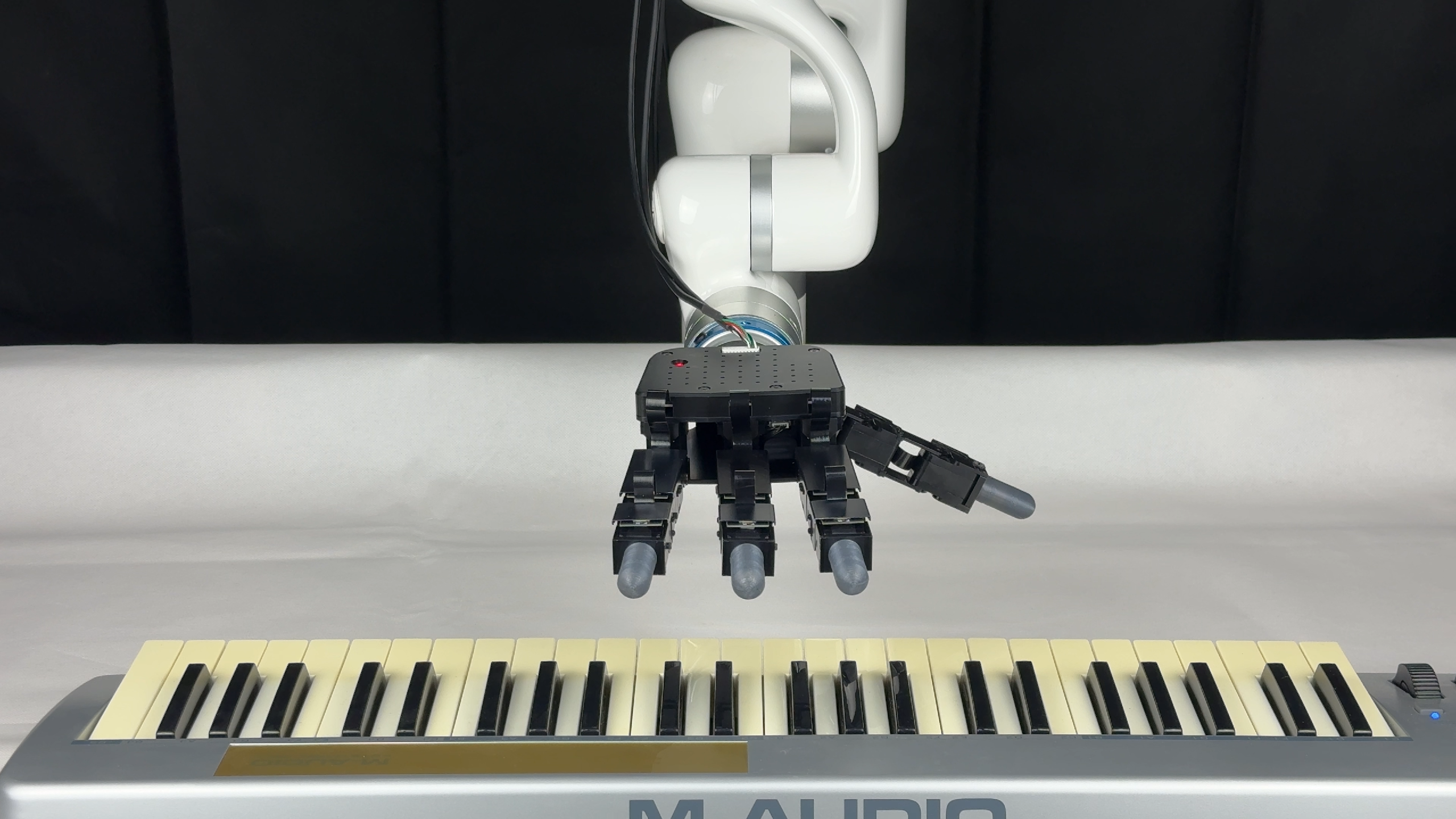}
  \caption{Real world hand and piano}\label{fig:real world env}
\end{subfigure}
\caption{Comparison of the simulated training environment to the real world setup. 
The real world environment consists of a multi-finger Allegro hand v4.0 mounted on an Ufactory xArm7 robot arm and an M-Audio Keystation 49e MIDI keyboard. 
The fingertips of the Allegro hand are replaced with smaller 3D-printed fingertips to allow pressing a single key at a time.}
\end{figure*}

Overall, our contributions are:
\begin{itemize}
    \item We introduce the first learning-based system capable of real world piano playing.
    \item Our codebase is open-sourced and additional videos are provided on our website. 
    This reduces the initial costs of research in this area and allows researchers to rapidly verify new approaches.
    \item We demonstrate that a Sim2Real2Sim is a promising path to explore for future piano playing contributions. 
    \item Our results can be used as a baseline to evaluate alternative approaches.
    \item We provide several important insights into the nature and challenges of playing piano in the real world.
\end{itemize}

%% file: 2_related.tex
\textbf{Musical Robots.}
Understanding the role of robotic systems in playing musical instruments goes a long way back to when highly sophisticated robots such as Takanishi's Anthropomorphic Flutist Robot~\cite{takanishi1996} were utilized to play the flute.
Similarly, early works concerning robots playing the piano indicate the extensive effort made by researchers who built humanoid-like robots that leveraged precise finger-arm movement for rendering tunes~\cite{Kato1987}.  
More recent works explored the application of robotic hands that are either specifically designed to play the piano~\cite{li2013, jen-chang2014} or utilize general purpose hands~\cite{scholz2019}.
Although path paving, these works heavily relied upon pre-programmed movements, limiting the robot's ability to adjust to musical styles or improvise on existing scores. 
This stands in contrast to the contributions utilizing machine learning algorithms to play piano.

\textbf{Learned Piano Playing.}
Playing the piano is a challenging benchmark for dexterity that requires precise and flowing movements.
The first RL based approach of playing piano was published in 2022~\cite{Xu2022}. 
The contribution showed that it is possible to play simple notes with correct velocity in simulation by utilizing vision based tactile sensors.
With the introduction of RoboPianist~\cite{Zakka2023} in 2023, it was shown that a RL-agent can be trained to play advanced bi-manual music. 
However, the policy needed to be trained for each song individually and manual fingering annotations are required.
Those limitations motivate the contribution of Zhao et al.~\cite{Zhao2024} in 2024.
They proposed fingering as an optimal transport problem~\cite{gabriel2019, villani2006} enabling the agent to efficiently discover fingering during training and removing the need for manual finger annotations while also being applicable to other morphologies.
To further improve the multi-song performance multiple contributions utilized imitation learning successfully~\cite{Zhao2024, qian2024, chen2025}. However, unlike prior works, which focused entirely on developing paradigms to play in simulation, we focused on translating the learned knowledge into the real world.

\textbf{Sim2Real Transfer.}
The goal of a Sim2Real transfer is to deploy a policy trained in the simulation in the real world.
Unfortunately, the naive approach without Sim2Real transfer techniques leads to a substantial drop in performance. 
This drop in performance is referred to as the Sim2Real gap.
Over time there were multiple methods proposed to minimize the Sim2Real gap \eg{} residual physics learning (RPL)~\cite{sontakke2023, zeng2020}, rapid motor adaptation (RMA)~\cite{Qi2022}, domain randomization (DR)~\cite{Handa2024, tobin2017, wang2024pens} and Sim2Real2Sim~\cite{chang2020, lim2022}.
They function by either reducing the differences between the simulator and the real world (Sim2Real2Sim, RPL) or by making the model more robust to variations of the environment (DR, RMA).
Our work merges both approaches by integrating Sim2Real2Sim and DR.

%% file: 3_approach.tex
\begin{table}[t]
    \centering
    \begin{tabular}{|l|c|c|} \hline 
         \textbf{Observations}&  \textbf{Units}& \textbf{Dimension}\\ \hline 
         Used joints of the Allegro hand&  rad& $12$\\ 
         Slider for parallel movement&  m& $1$\\
         Previous joint positions& mixed& $12+1=13$\\
         Currently pressed keys& one-hot-encoded&$49$\\ 
         Music sheet& one-hot-encoded&$6*49 = 294$\\ 
 \hline 
 \textbf{Total observation space} & &$\boldsymbol{369}$\\ \hline
    \end{tabular}
    \caption{The observation space of the agent.}
    \label{tab:observation space}
\end{table}

\textbf{Hardware Configuration.}
The hardware setup we used is shown in \fig{fig:real world env}. 
Our setup consists of an Ufactory xArm7 to move the end effector parallel to the piano, replicating the movement of the human forearm.
Attached to the end effector is an Allegro Hand v4.0. 
It consists of 4 fingers with 4 joints per finger. The thumb joints are not used in this contribution because they can not reliably reach the piano, leading to 12 active joints. 
The fingertips are replaced with 3D-printed fingers of approximately human dimensions, since the default fingertips are wider than a piano key and would prevent single-key presses.
Our piano is an M-Audio Keystation 49e MIDI Keyboard, which contains four octaves. 
Every keypress and every release triggers a MIDI event, which is forwarded to the host computer to allow for observation of the currently pressed keys.

\begin{figure}
    \centering
    \includegraphics[width=1\linewidth]{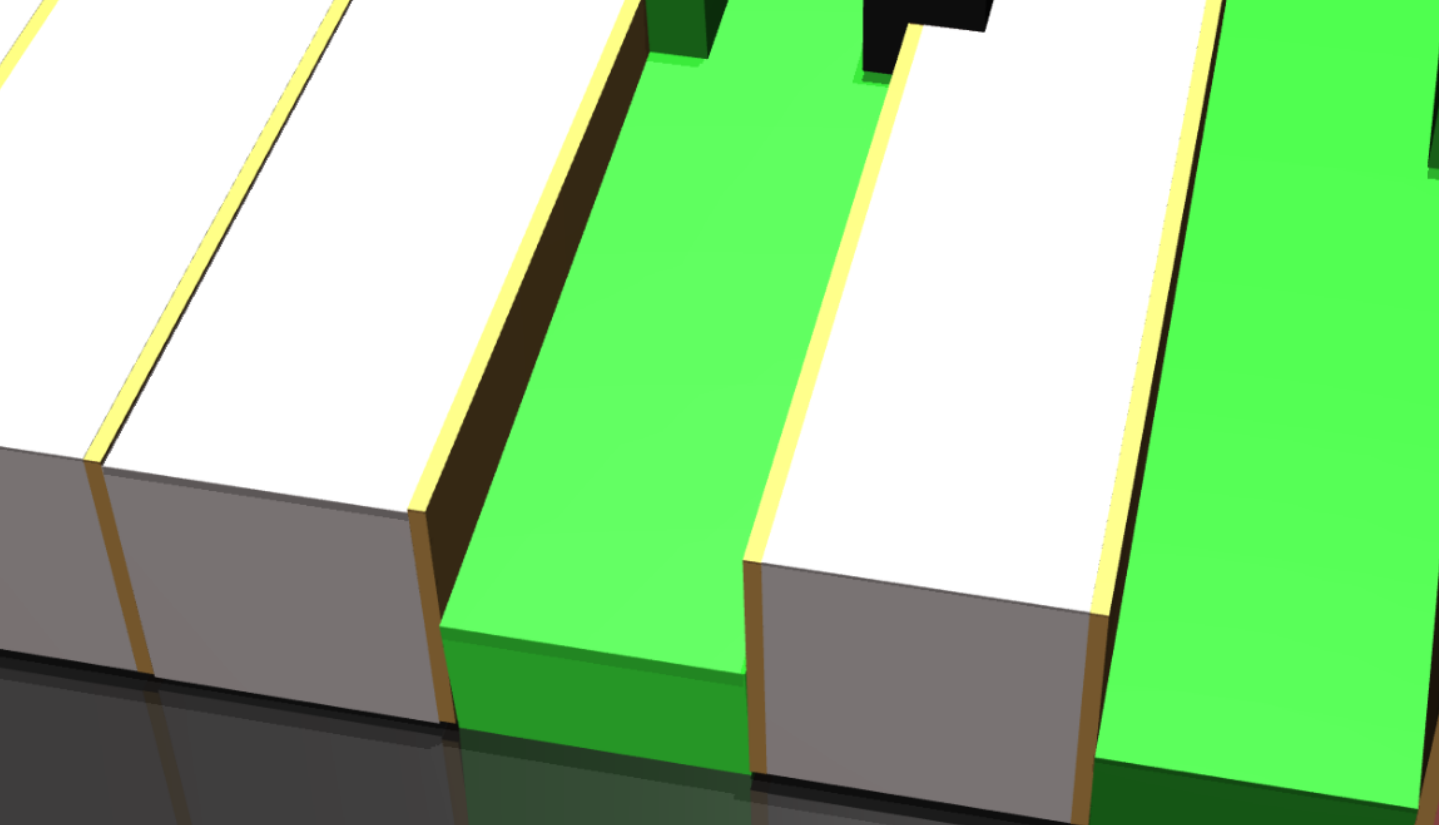}
    \caption{Rendering of the piano keys in our physics simulator. The yellow boundaries are fixed in place and simulate a wall, to prevent the agent from sliding to an adjacent key without lifting the finger. This is discussed in more detail in \sec{exp:fences}.}
    \label{fig:fences}
\end{figure}

\textbf{Simulation Environment.}
Our training environment was built upon the RoboPianist-Suite \cite{Zakka2023}, which is based on the Mujoco physics engine \cite{Todorov2012}. 
The scene of the simulation is displayed in \fig{fig:sim env}. 
It consists of the Allegro hand with custom fingertips and a model of the piano. 
Furthermore, we added additional boundaries between the white keys (see \fig{fig:fences}) to achieve an environment that mimics the real world more closely. 
These boundaries are discussed in more detail in \sec{exp:fences}.
All joints of the simulation use a PD controller with maximum jerk a low pass filter and damping.
This enables us to fine-tune the control parameters of all actuators individually during our Sim2Real2Sim workflow.

\textbf{Markov-Decision-Process.}
The action space consists of 13 continuous actions. 
One action per joint of the hand and one additional action controlling the forearm movement. 
The 369-dimensional observation space is shown in \tab{tab:observation space}. 
Note that the observations of the keys are discrete. 
This makes playing the piano more demanding because the agent cannot hover with the fingers on the keys to verify the correct finger positions, as described in the RoboPianist~\cite{Zakka2023}. 
Unfortunately, it is necessary because the MIDI standard used by the piano does not include partially pressed keys. 
Further, the observation space does not include any manual fingering annotations. 
Instead we encourage the agent to use efficient fingering by adding a component to the reward formulation.

\textbf{Reward Design.}
We will briefly discuss each component of our reward function 
$r_\text{total}=r_\text{energy}+r_{\Delta\text{-action}}+r_\text{keypress}+r_\text{optimal\_transport}+r_\text{collision}$
in this section.
The $r_\text{energy}$ reward penalizes fast and forceful movements. 
It avoids oscillations that don't lead to higher rewards and reduces the applied force to reduce the risk of damaging the real world setup.  
The $r_{\Delta\text{-action}}$ is a penalty proportional to the difference between consecutive actions to achieve smooth control signals.
The $r_\text{keypress}$ reward encourages the agent to press the correct keys.
To improve the fingering of our robot we added $r_\text{optimal\_transport}$, which is adopted from \cite{chen2025}.
However, instead of using the cartesian distance as costs we use the one dimensional difference in the horizontal axis to obtain the correct fingertip. 
Finally, the reward is calculated with respect to the cartesian distance.
The motivation for this design is to avoid reusing the same finger just because it has a lower height compared to another finger that is higher up but already above the correct key.
Lastly, we added $r_\text{collision}$ which is $0.5$ if there is no collision or only collisions between the piano and fingertips, otherwise it is $0$.
This leads to faster convergence as discussed in \sec{exp:coll-trick}.

\textbf{Evaluation Metrics.}
To quantify the success of the agent, we use the F1-score $F_1=2*(\text{recall}^{-1}+\text{precision}^{-1})^{-1}$, where recall addresses the question how reliable the agent presses the correct keys and precision addresses the question how reliable the agent avoids pressing wrong keys \cite{Zakka2023}. 

\begin{figure}[t]
    \centering
    \includegraphics[width=\linewidth]{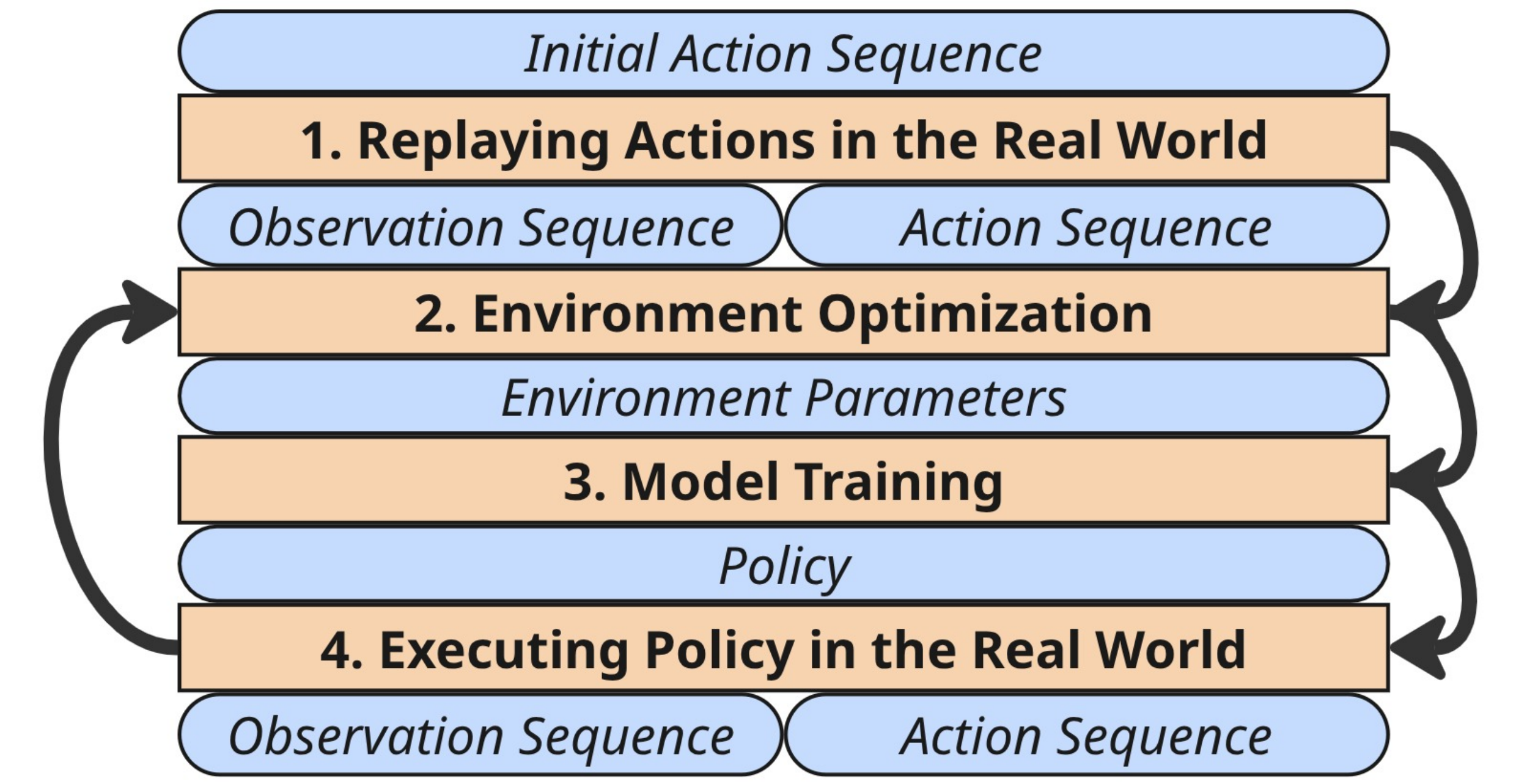}
    \caption{Diagram of our Sim2Real2Sim pipeline. 
    Real world observations from an initial action sequence are used to optimize the simulator. 
    Then a piano-playing policy is trained in simulation and deployed on the robot to generate new data for further simulation refinement.
    The process is explained in more detail in \sec{Sim2Real2Sim}.}
    \label{fig:Sim2Real2Sim}
\end{figure}

\textbf{Sim2Real2Sim.} \label{Sim2Real2Sim}
Our workflow is visualized in \fig{fig:Sim2Real2Sim}.
We start with an initial action sequence consisting of hand opening and closing motions. 
This sequence is executed on the real robot \textit{(step 1)} to collect observations used to optimize the simulation environment using Bayesian Optimization \textit{(step 2)}. The updated environment is then used to train a piano-playing policy \textit{(step 3)}, which is executed on the real robot \textit{(step 4)} to generate new observations and actions, enabling iterative refinement of the simulation \textit{(returning to step 2)}.

\textbf{Domain Randomization.}
To further decrease the Sim2Real gap, we utilize domain randomization of the height of the piano, the starting position of the hand, the dampening of the joints, the applied force of each actuator, the threshold of a key to be considered pressed, the stiffness of the springs holding the keys and the friction between the keys and the fingertips.

\textbf{Model Training.}
Our agent consists of three hidden layers with 128 nodes each.
The training uses DroQ~\cite{Hiraoka2022}, an off-policy soft actor-critic algorithm.
A new policy needs to be trained for each song, which takes about \SI{3}{\hour} \SI{20}{\minute} for $3e6$ epochs. 

%% file: 4_result.tex
\begin{figure*}[t]
\begin{subfigure}[t]{0.49\linewidth}
    \centering
    \includegraphics[width=\linewidth]{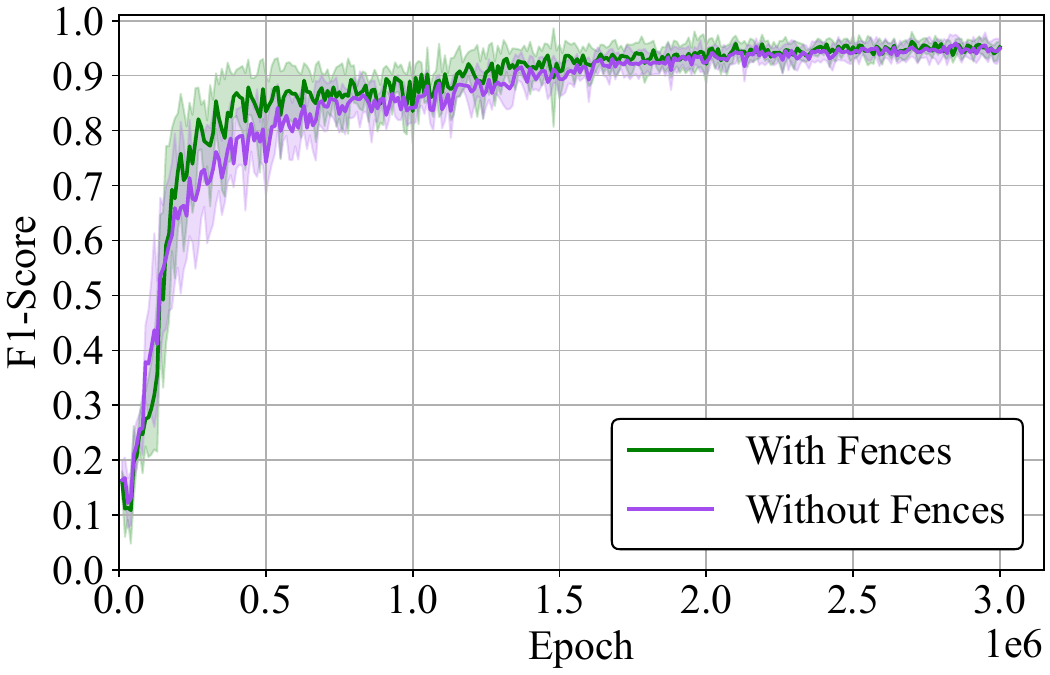}
\end{subfigure}
\hfill
\begin{subfigure}[t]{0.49\linewidth}
    \centering
    \includegraphics[width=\linewidth]{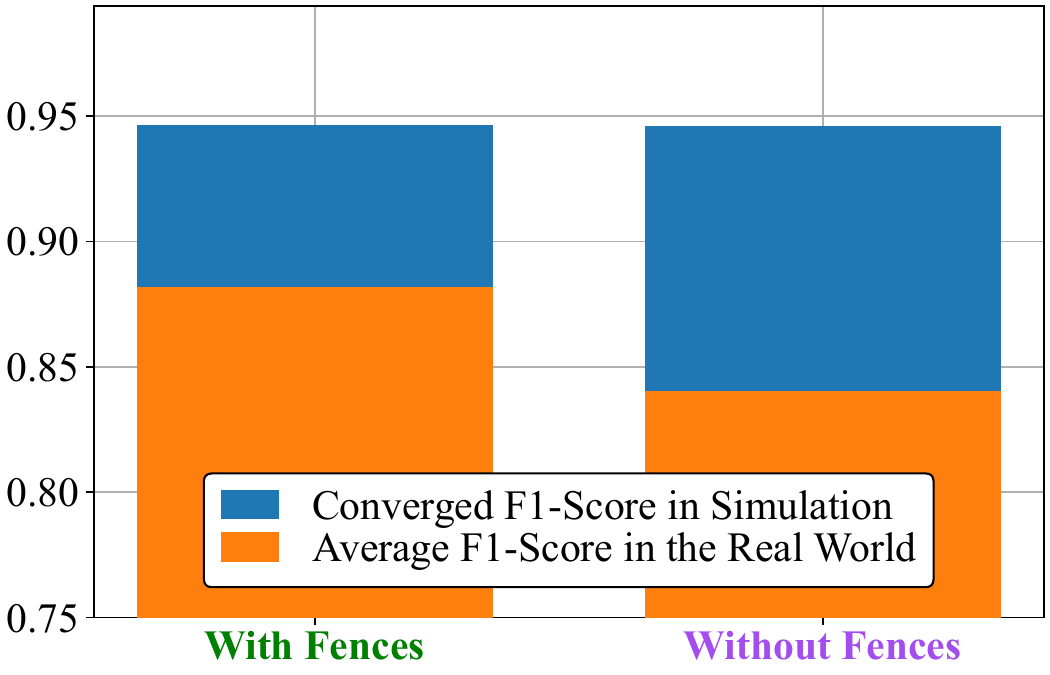}
\end{subfigure}
\caption{We analyze the effect of fixed boundaries, referred to as fences, between keys on the training convergence and on the Sim2Real gap. 
The environment with fences learns slightly faster, but ultimately both environments converge to similar performance. 
However, the real world evaluation shows an improved F1-score and therefore a smaller Sim2Real gap for the policy trained with fences compared to the fence-free environment.
This leads us to the conclusion that adding fences is beneficial, since it reduces the differences between the training environment and the real world.}
\label{fig:exp:fences}
\end{figure*}

In the following, we present a series of experiments designed to analyze key design choices of our approach and to evaluate its real world applicability. 
First, we investigate whether adding physical constraints in simulation, in the form of fences between piano keys, improves training efficiency and Sim2Real transfer. 
Second, we study the influence of a fingertip collision penalty in the reward formulation on the training and the resulting control strategy. 
Finally, we assess the real world performance and generalization capability of our approach across multiple songs by comparing the performance achieved in simulation and in the real world.

\subsection{\textbf{Is it beneficial to use fences between the white keys? (\fig{fig:exp:fences})}}\label{exp:fences}
\textbf{Setup.}
In this experiment we compare the training progression and the Sim2Real gap of two simulation environments. The first environment includes narrow fixed boundaries between the white keys, referred to as "fences" (see \fig{fig:fences}). The second environment does not have fences. We compare the convergence during training and the achieved F1-score in the real world.

\textbf{Results.}
During training the environment with fences progressed slightly faster compared to the environment without fences. Ultimately both environments converged to an F1-score of $0.946$. 
In the real world the model without fences achieves an F1-score of $0.840$, leading to a Sim2Real gap of $11.2\%$. 
In contrast, the model with fences achieved an F1-score of $0.881$, leading to an almost halved Sim2Real gap of $6.8\%$.

\textbf{Discussion.}
The results indicate that, in our setup, it is beneficial for the training and real world performance to add fences between the keys.
A possible explanation for the faster progression during training is that adding fences makes it simpler to press one key per finger and to hold it pressed without triggering the neighboring keys.
Our understanding of the reduced Sim2Real gap is, that the environment with fences makes the simulation behave more similar to the real world.
In the real world, applying lateral force to the fingertip while pressing a key does not activate adjacent keys. 
However, in our fence-free simulation, soft-body physics can cause unintended neighboring key presses. 
Adding fences forces the agent to completely lift its finger during transitions, better reflecting real world behavior and reducing the risk of getting stuck between keys.

% --------------------------------------------------------------

\begin{figure}[t]
    \centering
    \includegraphics[width=\linewidth]{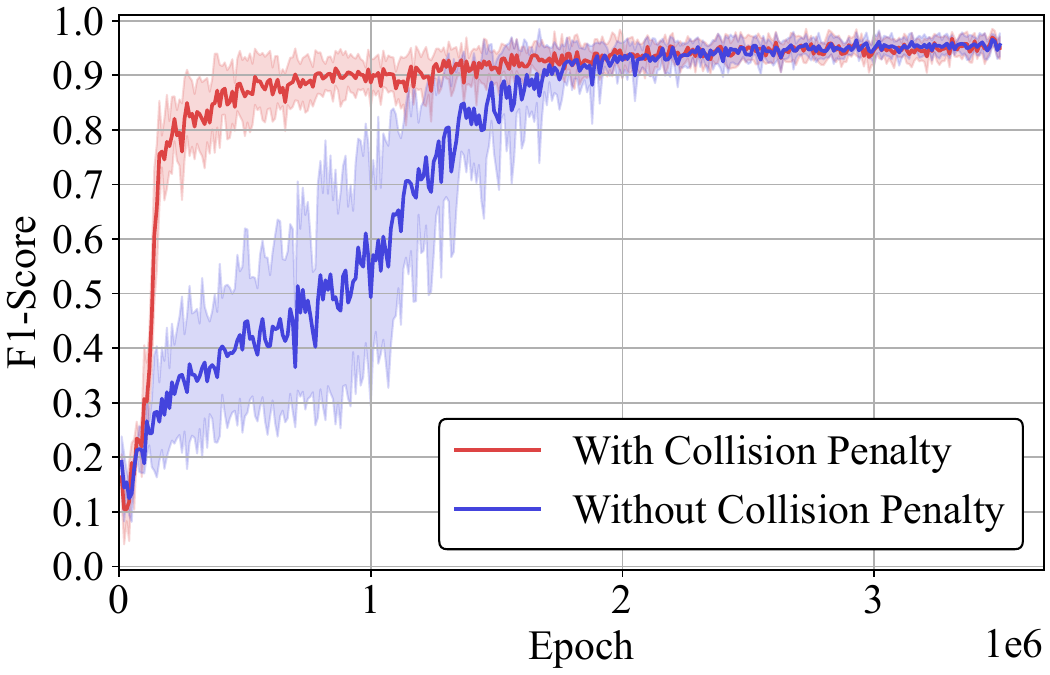}
    \caption{
The diagram compares the training progression of the F1-score with and without the fingertip collision penalty $r_\text{collision}$ in the reward formulation. 
Including the penalty leads to a rapid initial performance increase, while omitting it results in slower, near-linear improvement. 
Both reward formulations eventually converge to similar final performance.}
    \label{fig:exp:coll-trick}
\end{figure}

\subsection{\textbf{How does a fingertip collision penalty affect the training? (\fig{fig:exp:coll-trick})}}\label{exp:coll-trick}
\textbf{Setup.}
In this experiment we compare the reward formulation introduced in \sec{sec:approach} with a variation that does not include $r_\text{collision}$. We observe the progression of the F1-score during training.

\textbf{Results.}
The F1-score of the models trained with the collision penalty show a rapid initial performance increase before converging.
In contrast, the F1-score of the models without the penalty increases linearly at a slower rate.
The gap between both reward formulations is closed at about $\approx 2e6$ episodes and both rewards ultimately converge to the same value.

\textbf{Discussion.}
The component $r_\text{collision}$ is maximized by playing exclusively with the fingertips.
Both formulations converging to this strategy independently indicates that $r_\text{collision}$ does not have an influence on the learned strategy.
However, the models without the collision penalty require significant more time learning to use the fingertips properly instead of forcing the fingertips inside of the piano by exploiting the soft body physics.
This enables us to conclude that the collision penalty primarily accelerates learning rather than altering the final strategy. 
By discouraging unrealistic collisions early in training, it guides the policy towards a physically plausible fingertip usage. 
Once this behavior is learned, the reward term becomes constant and no longer influences the optimal policy, explaining the similar performance.

% --------------------------------------------------------------

\begin{figure}[t]
    \centering
    \includegraphics[width=1\linewidth]{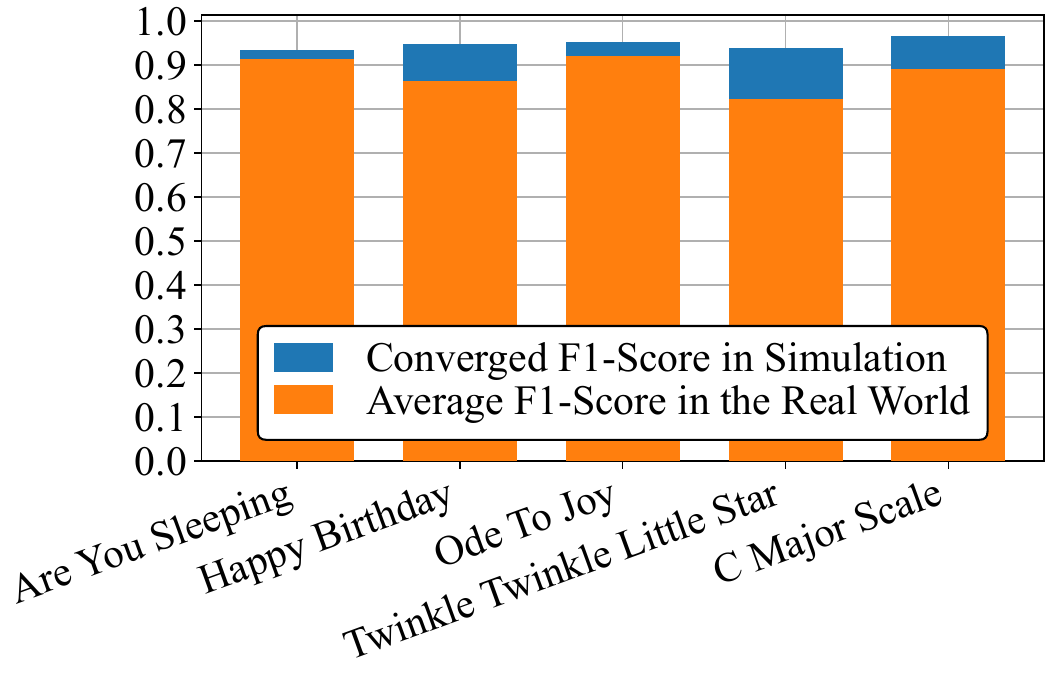}
    \caption{This experiment is an evaluation of the Sim2Real performance across all 5 songs. All songs are played successfully with an F1-score of $\ge 0.821$ in the real world. However, the Sim2Real gap differs significantly due to the inherent challenges of each song.}
    \label{fig:exp:rw-songs}
\end{figure}

\subsection{\textbf{How does our approach perform in the real world? (\fig{fig:exp:rw-songs})}}
\textbf{Setup.}
To evaluate the generalizability of our approach across songs we tested 5 simple songs: Are You Sleeping, Happy Birthday, Ode To Joy, Twinkle Twinkle Little Star and the C-Major Scale.
For each song we obtain the performance achieved in simulation and in the real world to calculate the Sim2Real gap.

\textbf{Results.}
All songs reached an F1-score between $0.937$ and $0.964$ in simulation and between $0.821$ and $0.919$ in the real world, however the Sim2Real gap differs significantly from $2\%$ up to $12.3\%$.
The average F1-score reached in simulation is $0.946$ and in the real world $0.881$.
Videos showcasing the performance of each song are displayed on our website \website{}.

\textbf{Discussion.}
The results show that the agent was able to play all 5 songs successfully. 
The differences in the Sim2Real gap are based on the characteristics of each song \eg{} Happy Birthday containing a black key and a faster note sequence in Twinkle Twinkle Little Star.
This also indicates that, although our method generalizes to simple songs, more advanced songs continue to pose challenges for the agent.

% sim [0.9325769036812908, 0.9464670658682636, 0.9517599596570852, 0.9372007366482504, 0.9642135642135642]
% rw [0.9130105900151286, 0.8627744510978044, 0.9199445284921836, 0.8218232044198897, 0.8914141414141413]
% gap [0.020980911696317217, 0.08842633599055227, 0.033427999194633684, 0.12310866574965595, 0.07550134690212518]

%% file: 5_limitations.tex
In this work, we share our findings on how to teach a robot to play simple songs on a real piano. 
We demonstrated a proof-of-concept that utilized Sim2Real2Sim to play piano in the real world. 
However, our system possesses several limitations that could be investigated in future research:

\textbf{Vision Modality.} 
With the current modalities the agent almost entirely relies on its proprioception, which makes perceiving the real world hard.
The agent needs to learn the position of each key during training and blindly trust into this knowledge while playing in the real world.
To perceive an inaccurate alignment of the fingertip, the agent needs to make a wrong keypress.
An additional vision modality would allow the robot to correct itself before pressing the key, potentially improving the overall accuracy.

\textbf{Tactile Sensors}. 
Another limitation of our work is the lack of additional sensing modalities, including touch sensing. 
Prior works~\cite{Xu2022, nakahara2025} indicate that touch can enable the robot to perceive additional information about the environment.
We considered integrating existing touch sensing solutions, such as DIGIT~\cite{Lambeta2020}.
However, these sensors are larger than a human finger, and resulted in pressing two keys at once.
We believe that integrating future generations of tactile sensors that are smaller would be a valuable research topic.

\textbf{Advanced Fingering}. 
Our work focuses on the Allegro hand which does not allow for finger abduction.
This significantly limits the keys each finger can reach without moving the whole hand and it prevents the thumb from reaching the piano.
Using a robotic hand that allows for abduction would enable to agent to use more advanced fingering.
Further, in human piano playing, the wrist takes a key role to extend the range of each finger for certain fingering.
In our work the orientation of the wrist is fixed further reducing the range of each finger.
These hardware limitations resulted in movement of the whole hand at the level of the arm.
Future research could investigate ways to minimize these movements for more natural results.

\textbf{Advanced Songs.} 
This work focuses on five simple songs.
Playing more advanced songs with faster melodies that require rapid movements is the natural continuation in this direction.
This also mimics the progression of a young pianist that learns to play piano with simple songs and one hand before focusing on advanced songs.

\textbf{Bimanual Playing}. 
Many songs require the use of two hands.
This also introduces the challenge of coordinating twice the number of joints and enables more complex fingering techniques.
However, currently our hardware setup is limited to a single hand.
In the future, we hope to extend to the bimanual case and investigate how to learn more complex piano movements, thus being able to increase the overall difficulty of the played songs.

%% file: 6_conclusion.tex
In this work, we showcase the first learned approach for playing piano with a real world robotic hand.
Specifically, we employ Sim2Real2Sim where we iterate in the real world before training the final deployed policy.
To mimic the real world more accurately, we further introduced boundaries between the white keys in simulation.
These boundaries can be easily adopted by other researchers to increase the accurately of their simulations to the real world.
Our experimental results demonstrate that our model can play several simple pieces successfully in the real world. 
Nonetheless, there is still a long way before we can achieve robots with the same level of proficiency and elegance as human pianists.
Beyond the artistic \textit{raison d'être}, we believe that playing piano is a relevant and challenging benchmark for dexterous contact-rich manipulation, and advances in this task can be more widely relevant for the whole manipulation community.

%% file: 99_acknowledgments.tex
We thank Ken Nakahara for the support and tips regarding operating the robotics and throughout the project. 
This work was supported by the Robotics Institute Germany (RIG), by the German Research Foundation (DFG) under the Cluster of Excellence CARE: Climate-Neutral And Resource-Efficient Construction (EXC 3115) project number 533767731, by the Bundesministerium für Bildung und Forschung (BMBF) and German Academic Exchange Service (DAAD) in project 57616814 (\href{https://secai.org/}{SECAI}, \href{https://secai.org/}{School of Embedded and Composite AI}), by the project "Genius Robot" (01IS24083) funded by the Federal Ministry of Education and Research (BMBF), .

%% file: appendix/appendix_main.tex
\begin{appendices}

\input{appendix/0_prologue}

\section{Our Hybrid Execution Setup}
\input{appendix/1_approach}

\section{Preliminary Results}
\input{appendix/2_result}

\section{Conclusion of Hybrid Execution} 
\label{app:approach}
\input{appendix/3_conclusion}

\end{appendices}

%% file: appendix/0_prologue.tex
During the development of our system we also experimented with Hybrid Execution as another Sim2Real transfer technique.
Although this was path paving for us, we eventually focused on Sim2Real2Sim in this contribution.
In the future we consider combining our Sim2Real2Sim approach with Hybrid Execution to further improve our results.
In this section we will describe our preliminary results using Hybrid Execution.

%% file: appendix/1_approach.tex
\textbf{Hardware Configuration Differences.}
The hardware setup is consistent with the configuration described in \textbackslash{}sec\{sec:approach\}.
The only difference is the use of a right-handed Allegro hand for organizational reasons.
Since the thumb is disabled and the remaining fingers are symmetrical, this does not introduce a significant difference.

\textbf{Simulation Environment Differences.}
The simulation environment does not use fences (see \fig{fig:fences}). 
Further, instead of using a pd controllers, we use mujoco position actuators with joint damping to actuate the joints.
Besides those differences the environment is analogous.

\textbf{Markov-Decision-Process.}
The action space still consists of 13 continuous actions. 
One action per joint of the hand and one additional action for the wrist movement, parallel to the piano. 
The observation space however is only 356-dimensional, since it does not include the previous joint positions (compare table \ref{tab:observation space}).

\textbf{Reward Design.}
The reward function used in the following experiments
$r_\text{total}=r_\text{energy}+r_\text{hand\_position}+r_\text{keypress}+r_\text{sliding}$
does not utilize optimal transport to explicitly encourage fingering.
Instead, the reward $r_\text{hand\_position}$ is designed to guide the wrist towards intended keys, while $r_\text{keypress}$ encourages the agent to press the correct keys. 
The $r_\text{sliding}$ reward is designed to penalize fast sideways movement of the hand while keys are pressed to avoid damaging the robot.
The $r_\text{energy}$ reward penalizes fast movements that need a lot of force. 
Further, it avoids oscillations that don't lead to higher rewards.  

\begin{figure*}[t]
    \centering
    \includegraphics[width=1\textwidth]{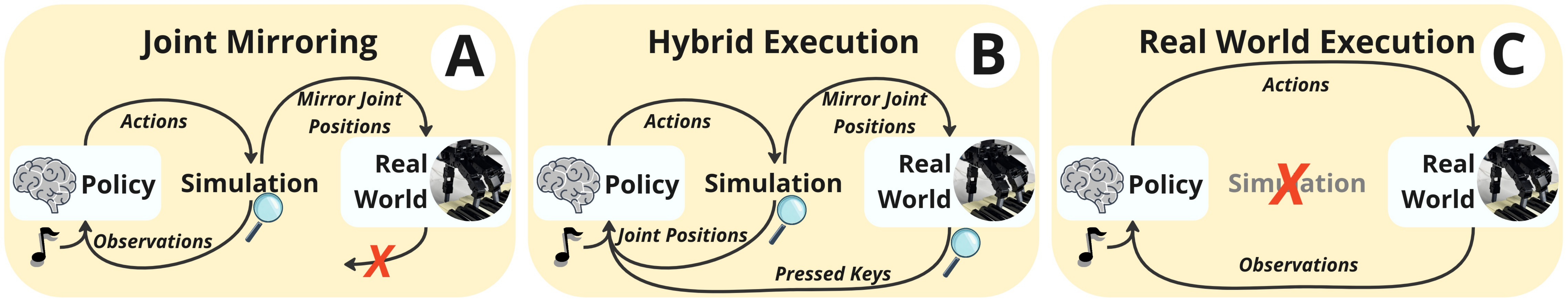}
    \caption{The diagram compares the three execution modes:
    A) In joint mirroring, the whole observation space is obtained from the simulated environment. 
    B) In hybrid execution, only the pressed keys are based on the real world, while everything else is simulated.
    C) In real world execution, all observations are based on the real world.}
    \label{fig:execution_modes}
\end{figure*}

\textbf{Model Training.}
During training we use domain randomization (DR) as introduced in \sec{sec:approach}. 
However we don't apply our Sim2Real2Sim procedure.
Instead, we focus on three execution modes.

\textbf{Execution Modes.} To deploy the trained policy in the real world, we introduce three modes of execution, which are also displayed in \fig{fig:execution_modes}:

\underline{Joint Mirroring:} In this mode, all observations are based on the simulation.
The real world robot mirrors the joint positions of the simulation into the real world.
This leads to a deterministic sequence of actions, but it is impossible to adapt to the environment.

\underline{Real World Execution:} This mode is the opposite of joint mirroring and the most intuitive approach. 
All observations are based on the real world and the actions are applied directly to the robot. No simulation is used in this mode.

\underline{Hybrid Execution:} The observation space of the model is split into observations of the real world and observations of the simulation. 
This enables the model to be deployed in the real world while being less robust compared to the real world execution.
Hybrid execution comes with the risk of a potentially diverging simulation, which needs to be tested carefully.
In this work, the pressed keys are based on the real world, while the joint positions are based on the simulation.\label{Hybrid Execution}

\textbf{Song Collection.}
We validate our models on multiple songs with different characteristics:

\underline{Twinkle Twinkle Little Star}: 
This song is a simple children's song. 
Its most relevant characteristic is the big interval at the beginning of the song and the memorable rhythm of the song.
It further serves as an everyday life example.

\underline{C-Major-Scale}: 
Having a scale in the repertoire is useful since it evaluates how the model changes from one key to another. 
For humans, it is intuitive to use all the fingers for playing the scale. 
We will discuss in the experiments whether our agent exploits the same strategy.

\underline{D-Major-Scale}: 
The major difference between the C-Major scale and the D-Major scale is that the D-Major scale contains black keys as well.
This allows us to observe the fingering technique of the agent while pressing the black keys.

\underline{Chord Progression}: 
In the first three songs, the agent only needs to press one key at a time. 
In this song, we train on a series of four chords instead, which lets us observe the multi-finger performance.

%% file: appendix/2_result.tex
\begin{figure*}
\begin{subfigure}[t]{0.49\linewidth}
    \centering
    \includegraphics[width=\linewidth]{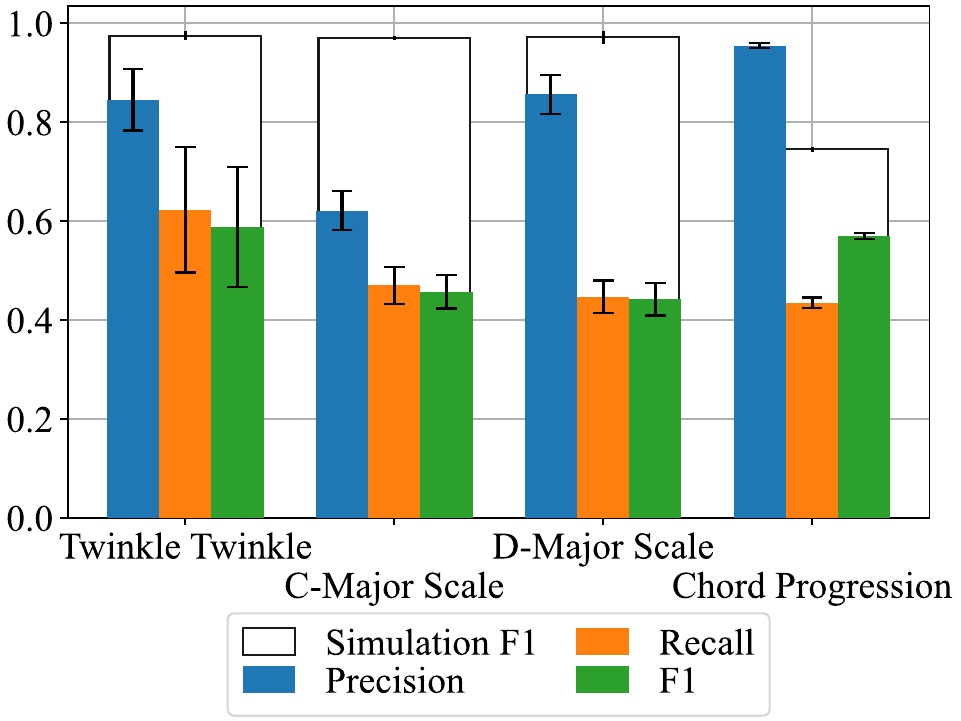}
    \caption{
    Comparison of several songs in the real world using hybrid execution.
    % For each song, we compare the precision, recall, and F1 score.
    % 
    % The displayed data shows that although we observe a significant Sim2Real gap, we are able to play all four songs successfully.
    % This indicates that our approach is generalizable and can be applied to other applications as well.
    }
    \label{fig:model_performance}
\end{subfigure}
\hfill
\begin{subfigure}[t]{0.49\linewidth}
    \centering
    \includegraphics[width=\linewidth]{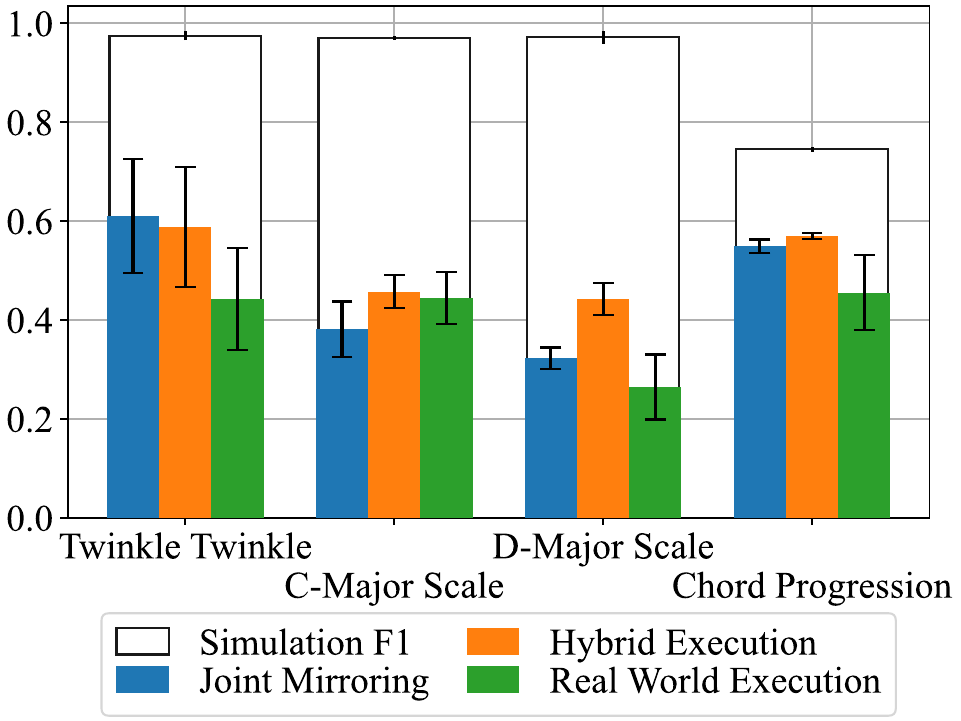}
    \caption{
    Comparison of the three execution modes across multiple songs.
    % The diagram compares our execution modes on multiple songs. 
    % The best performing mode is hybrid execution. 
    % It allows the agent to correct itself while the joints behave similarly to what the agent was trained on in simulation.
    }
    \label{fig:exp:modes}
\end{subfigure}
\caption{
As a reference is in both diagrams the F1 score reached in the simulation is provided. 
The results show that, with our setup, hybrid execution is the best-performing execution mode.
This indicates that the simulation does not significantly diverge from the real world, which enables us to improve upon the performance of the execution mode real world execution.
The results also show the potential of hybrid execution in settings where it is hard to achieve a model robust enough for a complete real world execution.
}
\end{figure*}

We provide videos showcasing real world runs for each of the experiments on our website: \website.
In this section, we experimentally answer the following questions:

\subsection{\textbf{How does our model perform? (\fig{fig:model_performance})}}
\textbf{Setup.}
To evaluate the performance of our approach, we trained on four different songs. 
Each of those songs was trained twice on independent seeds, which led to eight models in total.
Since every run in the real world differs, we executed each model three times. 
The figure shows the average and standard deviation of the precision, recall and the resulting F1 score for each song after using hybrid execution as introduced in \app{Hybrid Execution}.  
Other execution modes are evaluated in \app{exp: modes}.
As a reference for the Sim2Real gap is also the performance in simulation provided.

\textbf{Results.}
All four songs have in common that the precision is higher than the recall score. 
A possible explanation is that recovering from a wrong keypress consists of multiple steps.
For the precision, just lifting the finger is sufficient since no wrong keys are pressed anymore.
However, to improve the recall score, the agent needs to lift the finger, move it sideways, and press the correct key.
This sequence of actions is significantly more demanding and takes more time.
By the time this takes, a new key might be already targeted, resulting in skipping some keys.
Another explanation for the higher recall compared to the precision score is that maximizing the recall score is a simple task compared to maximizing the precision score. 
To maximize the precision, the agent needs to press at least all the correct keys. 
But to maximize the recall, the agent can easily press no keys at all.
Another observation is that the Chord Progression performed the worst in the simulation, with an F1 score of  $\approx0.2$ less than all other songs.
This might be because playing chords requires the robot to press multiple keys at once. 
Therefore, the robot needs to organize multiple fingers at the same time, which might require a bigger model than just using one finger with occasional finger switches.
Further, playing with multiple fingers simultaneously complicates the exploration of fingering techniques.
However, the observed F1 score of the Chord Progression is significantly above the average, with almost no variation.
This results in the smallest Sim2Real gap of this experiment.
A possible reason is that each chord needs to be pressed for a longer duration than a single key in the other songs. 
This enables the robot to correct itself more easily, decreasing the Sim2Real gap.

\textbf{Discussion.}
In this experiment, we demonstrated that it is possible to train a piano-playing model using common RL techniques that can be deployed using hybrid execution.
Our models are able to play all four songs successfully in simulation and in the real world.
This demonstrates effectively that playing piano in the real world with a learned model is within the reach of our current capabilities.

\subsection{\textbf{Which execution mode has the most potential? (\fig{fig:exp:modes})}}\label{exp: modes}
\textbf{Setup.}
In this experiment we perform an ablation study over the execution modes introduced in \app{app:approach}: joint mirroring, hybrid execution, and real world execution. 
No DR was used for the joint mirroring and hybrid execution. 
For the real world execution we decided to use a moderate intensity of DR with $C_{dr}=0.5$. 
The effect of DR is further discussed in \app{exp: dr}.
Analog to the prior experiment, we trained each model configuration twice and executed each model 3 times per execution mode. 
The figure shows the average and standard deviation of 6 runs for each mode.

\textbf{Results.}
Results show that the agent's performance differs across the execution modes depending on the song.
While joint mirroring performed among the best for Twinkle Twinkle Little Star and the Chord Progression, it performed worse than other modes for both scales. 
The real world execution, on the other side, performed well for the C-Major Scale but underperformed for all other songs.
The hybrid execution was the best-performing mode for three out of four songs. 
The only exception is Twinkle Twinkle Little Star, where the hybrid execution mode was slightly behind joint mirroring.
To understand the superiority of hybrid execution, we need to discuss the advantages and disadvantages of each execution mode separately.
When using joint mirroring, we don't observe the real world, which makes it impossible to correct wrong keypresses. 
Further, the simulation might diverge from the real world which would reduce the performance.
The advantage of this mode is that it does not need DR, which reduces the resource requirements and the number of hyperparameters.
The lack of adaptability is especially a problem for both scales since the model learned to slide sideways across the keys during the training.
Unfortunately, this strategy is hard to execute in the real world since the finger slips differently every time.
The real world execution has the advantage of observing the whole state of the real world robot. 
This ideally allows the robot to react in the best way possible since no information is hidden. 
Unfortunately, the additional information also requires a more robust model, which can be reached by utilizing DR as shown in \app{exp: dr}.
However, DR makes the training environment significantly more demanding.
In our experiments, the real world execution mode was not able to outperform the other execution modes.
The hybrid execution combines both modes.
Most of the observations are still generated in simulation, which makes it easier for the agent to transfer the learned knowledge into the real world.
However, by providing observations of the actually pressed keys, the robot is able to correct its behavior.
This way, the agent behaves similarly to joint mirroring when the pressed keys match the currently pressed keys in the simulation. 
However, as soon as the simulation differs from the real world, the agent is able to correct itself. 
This is possible while not using any DR at all, which makes the process even more favorable.

\textbf{Discussion.}
We showed that hybrid execution performed the best in our setup. 
It allows the agent to correct itself while being able to use the joints in the same way as they behaved during training.
This makes hybrid execution modes especially interesting for future work with the goal of achieving human-like proficiency in complicated settings, where it is hard to deploy a model entirely into the real world.

\begin{figure*}
\begin{subfigure}[t]{0.49\linewidth}
    \centering
    \includegraphics[width=\linewidth]{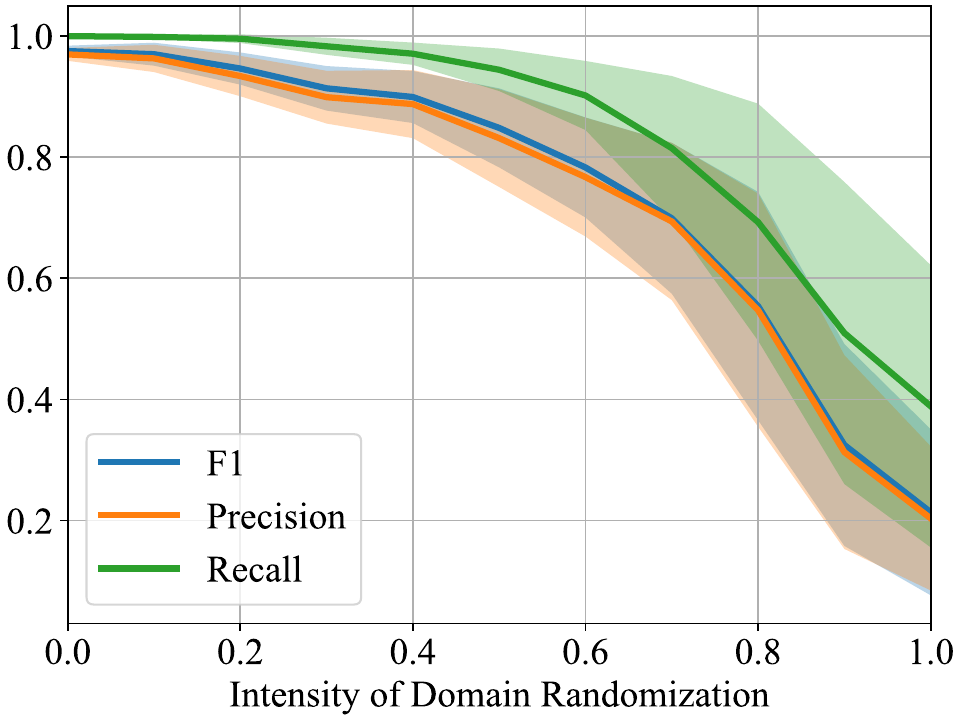}
    \caption{The diagram shows the effect of DR on the performance in simulation.}
    \label{fig:exp:dr:sim}
\end{subfigure}
\hfill
\begin{subfigure}[t]{0.49\linewidth}
    \centering
    \includegraphics[width=\linewidth]{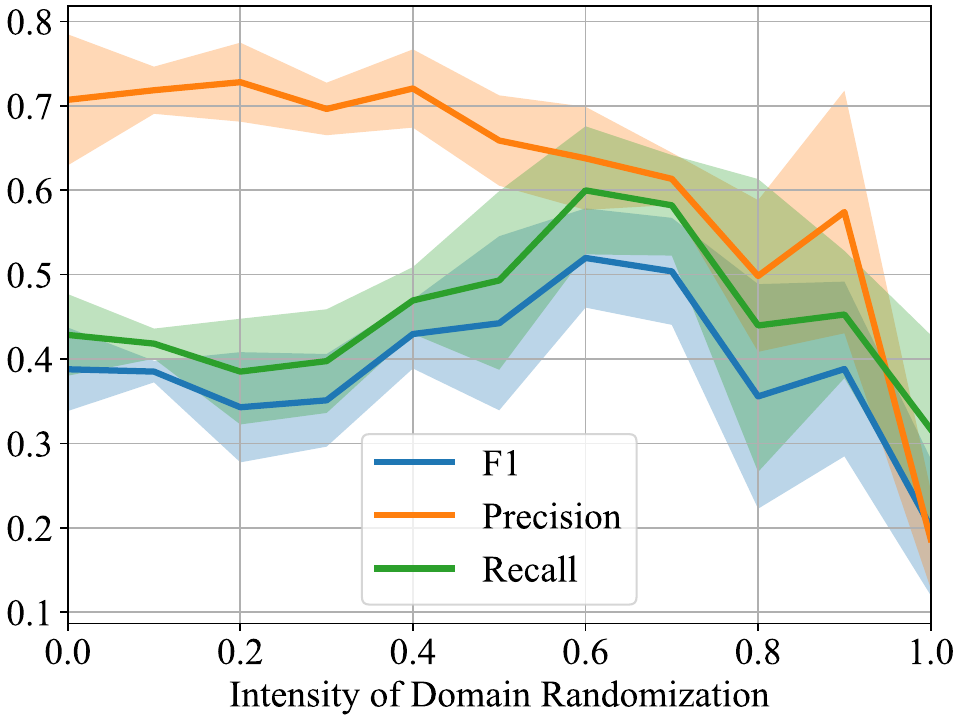}
    \caption{This diagram shows the effect of DR on the model performance in the real world using real world execution.}
    \label{fig:exp:dr:rw}
\end{subfigure}
\caption{
    The more DR is applied, the more demanding the simulation environment becomes. 
    This leads to a drop in performance, as observed in \fig{fig:exp:dr:sim}.
    However, DR leads to a more robust model, which can be observed in \fig{fig:exp:dr:rw}.
    With too much DR, the model is not able to play piano successfully, neither in simulation nor in the real world.
    }
\end{figure*}

\subsection{\textbf{How does the intensity of DR affect the real world performance? (\fig{fig:exp:dr:sim}, \fig{fig:exp:dr:rw})}}\label{exp: dr}
\textbf{Setup.}
During this experiment, we trained 11 different configurations with an increasing intensity of DR. 
To quantify the intensity of DR, we introduce the parameter $C_{dr}\in[0,1]$, with $C_{dr}=0$ representing no DR and $C_{dr}=1$ representing the maximum amount of DR.
The models are then evaluated using real world execution. 
The figure shows the average and standard deviation of multiple runs. 
To reduce the tear on the robot and since the most interesting runs are the best performing runs, we decided to reduce the number of runs for higher and lower intensities of DR.
We calculate the average of at least 3 runs for each configuration.
Further, for $C_{dr}\in[0.2,0.8]$ we evaluated at least 5 runs and for $C_{dr}\in[0.3,0.7]$ we evaluated 7 runs.
The models are evenly distributed across 4 different seeds.

\textbf{Results.}
As visualized in \fig{fig:exp:dr:sim}, the performance in simulation decreases the more DR is applied during training.
This is because the environment behaves differently for every iteration when DR is used, which requires a more robust model.
When the number of iterations during the training stays the same, but the environment gets more demanding, a performance decrease is to be expected.
Further, we observe that the recall in the simulation is higher than the precision.
A possible explanation is that our physics framework is based on soft-body physics.
This introduces a tendency to press two neighboring keys at once during training, while only one key is intended to be pressed.
This effect reduces the precision while maintaining a high recall score.
The real world performance, on the other hand, is displayed in \fig{fig:exp:dr:rw}. 
In contrast to the observations in simulation, the recall is significantly lower than the precision score in the real world.
This is because, in the real world, it's difficult to press multiple keys using just one finger.
The plastic surfaces of the keys and fingers are slippery on each other, which causes the finger to slide onto one key.
Once a key is pressed, the robot needs to lift the finger to press another key because the body of the adjacent keys acts as a border.
This makes it harder to press the targeted key, since the agent might slide to the wrong side, while in simulation, the robot would have pressed both keys.
This ultimately reduces the recall score in the real world.
The progression of the precision in the real world (\fig{fig:exp:dr:rw}) is similar to the progression of the precision in the simulation (\fig{fig:exp:dr:sim}). 
However, the real world precision is decreased by $\approx0.2$ for all intensities.
The recall on the other side has a very different progression compared to the simulation.
Without DR, the recall is at $\approx0.44$.
Increasing the intensity of DR leads to a reduced recall score of $\approx0.35$ at $C_{dr}=0.2$. 
From there, the recall increases significantly until the maximum of $\approx0.6$ is reached at $C_{dr}=0.6$, which also leads to the highest F1 score of $\approx0.52$.
For $C_{dr}>0.6$ the recall drops rapidly.
A possible explanation for the similar progression of the precision is that an unintentionally pressed key can be corrected by just lifting the finger. 
This skill is simple to learn even without applying DR.
The recall on the other side, significantly benefits from the DR.
To correct a wrong keypress, the agent needs to move the finger to a different key.
This requires more precise controls compared to just moving the finger up, which can be achieved by making the model more robust with DR.
However, the decreased recall performance at $C_{dr}=0.2$ is harder to explain.
Our understanding is that the agent seldom ends up in those wrong states without DR.
Therefore, the agent is also not able to learn how to recover from those states.
For $C_{dr}>0.6$, a drop in performance is expected since the performance in simulation drops as well.

\textbf{Discussion.}
The best performance can be achieved with moderate DR. 
Too little DR leads to worse performance since the model is not able to recover successfully after the environment behaved in a different way compared to the training - the model is not robust enough. 
Too much DR on the other side worsens the performance since the model is not able to learn successfully during training.
As a consequence of these results, we recommend researchers to carefully tune the intensity of DR, since it significantly impacts the real world performance.

%% file: appendix/3_conclusion.tex
In the appendix, we present an alternative approach for learning to play piano in the real world.
Specifically, we adopt a Sim2Real paradigm, in which a policy is first trained in simulation and subsequently deployed in the real world environment.
We further introduce Hybrid Execution, a novel inference framework designed for real world deployment. 
Hybrid Execution enables the agent to both observe and act in the real world with a reduced Sim2Real gap. 
Hybrid Execution is particularly well-suited for complex tasks such as piano playing, where achieving sufficient robustness for direct real-world deployment remains challenging.
However, an important consideration for applying Hybrid Execution more broadly is the potential divergence between simulation and reality, which may lead to degraded performance.
Overall, our experimental results demonstrate that Hybrid Execution enables the robotic system to successfully perform several simple piano pieces in a real-world setting.